\documentclass[11pt,a4paper]{article}
\usepackage[hyperref]{acl2018}
\usepackage{times}
\usepackage{latexsym}
\usepackage{times}
\usepackage{amsmath}
\usepackage{amsfonts}
\usepackage{latexsym}
\usepackage{graphicx}

\usepackage{url}

\aclfinalcopy 
\def\aclpaperid{1193} 

\title{Personalized Language Model for Query Auto-Completion}

\author{\textbf{Aaron Jaech and  Mari Ostendorf} \\
  University of Washington \\
  {\tt \{ajaech, ostendor\}uw.edu}}

\date{}

\begin{document}
\maketitle
\begin{abstract}
Query auto-completion is a search engine feature whereby the system suggests completed queries as the user types. Recently, the use of a recurrent neural network language model was suggested as a method of generating query completions. We show how an adaptable language model can be used to generate personalized completions and how the model can use online updating to make predictions for users not seen during training. The personalized predictions are significantly better than a baseline that uses no user information.
\end{abstract}

\section{Introduction}

Query auto-completion (QAC) is a feature used by search engines that provides a list of suggested queries for the user as they are typing. For instance, if the user types the prefix ``mete" then the system might suggest ``meters" or ``meteorite" as completions. This feature can save the user time and reduce cognitive load \cite{cai2016survey}.

Most approaches to QAC are extensions of the Most Popular Completion (MPC) algorithm \cite{bar2011context}. MPC suggests completions based on the most popular queries in the training data that match the specified prefix. One way to improve MPC is to consider additional signals such as temporal information \cite{shokouhi2012time,whiting2014recent} or information gleaned from a users' past queries \cite{shokouhi2013learning}. This paper deals with the latter of those two signals, i.e. personalization. Personalization relies on the fact that query likelihoods are drastically different among different people depending on their needs and interests.

Recently, \newcite{park2017neural} suggested a significantly different approach to QAC. In their work, completions are generated from a character LSTM language model instead of by ranking completions retrieved from a database, as in the MPC algorithm. This approach is able to complete queries whose prefixes were not seen during training and has significant memory savings over having to store a large query database.

Building on this work, we consider the task of personalized QAC, advancing current methods by combining the obvious advantages of personalization with the effectiveness of a language model in handling rare and previously unseen prefixes. The model must learn how to extract information from a user's past queries and use it to adapt the generative model for that person's future queries. To do this, we leverage recent advances in context-adaptive neural language modeling. 
In particular, we make use of the recently introduced FactorCell model that uses an embedding vector to additively transform the weights of the language model's recurrent layer with a low-rank matrix \cite{jaech2017low}. 
By allowing a greater fraction of the weights to change during personalization, the FactorCell model has 
advantages over the traditional approach to adaptation of concatenating a context vector to the input of the LSTM \cite{mikolov2012context}.

\begin{table}[]
\centering
\begin{tabular}{lll}
 & \textbf{Cold Start} & \textbf{Warm Start} \\
\hline
1 & bank of america & bank of america \\
2 & barnes and noble & basketball \\
3 & babiesrus & baseball \\
4 & baby names & barnes and noble \\
5 & bank one & baltimore
\end{tabular}
\caption{Top five completions for the prefix ``ba" for a cold start model with no user knowledge and a warm model that has seen the queries espn, sports news, nascar, yankees, and nba. }
\label{table:example}
\end{table}

Table \ref{table:example} provides an anecdotal example from the trained FactorCell model to demonstrate the intended behavior. The table shows the top five completions for the prefix ``ba" in a cold start scenario and again after the user has completed five sports related queries. In the warm start scenario, the ``baby names" and ``babiesrus" completions no longer appear in the top five and have been replaced with ``basketball" and ``baseball".

The novel aspects of this work are the application of an adaptive language model to the task of QAC personalization and the demonstration of how RNN language models can be adapted to contexts (users) not seen during training. An additional contribution is showing that a richer adaptation framework gives added gains with added data.


\section{Model}
\label{sec:model}

Adaptation depends on learning an embedding for each user, which we discuss in Section \ref{sec:learning_embeddings}, and then using that embedding to adjust the weights of the recurrent layer, discussed in Section \ref{sec:recurrent_adaptation}. 

\subsection{Learning User Embeddings}
\label{sec:learning_embeddings}

During training, we learn an embedding for each of the users. We think of these embeddings as holding latent demographic factors for each user. Users who have less than 15 queries in the training data (around half the users but less than 13\% of the queries) are grouped together as a single entity, $user_1$, leaving $k$ users. The user embeddings matrix $\mathbf{U}_{k \times m}$, where $m$ is the user embedding size, is learned via back-propagation as part of the end-to-end model. The embedding for an individual user is the $i$th row of $\mathbf{U}$ and is denoted by $u_i$.

It is important to be able to apply the model to users that are not seen during training. This is done by online updating of the user embeddings during evaluation. When a new person, $user_{k+1}$ is seen, a new row is added to $\mathbf{U}$  and initialized to $u_1$. Each person's user embedding is updated via back-propagation every time they select a query. When doing online updating of the user embeddings, the rest of the model parameters (everything except $\mathbf{U}$) are frozen.

\subsection{Recurrent Layer Adaptation}
\label{sec:recurrent_adaptation}

We consider three model architectures which differ only in the method for adapting the recurrent layer. First is the unadapted LM, analogous to the model from \newcite{park2017neural}, which does no personalization. The second architecture was introduced by \newcite{mikolov2012context} and has been used multiple times for LM personalization \cite{wen2013recurrent,huang2014enriching,li2016persona}. It works by concatenating a user embedding to the character embedding at every step of the input to the recurrent layer. \newcite{jaech2017low} refer to this model as the ConcatCell and show that it is equivalent to adding a term $\mathbf{V}u$ to adjust the bias of the recurrent layer. The hidden state of a ConcatCell with embedding size $e$ and hidden state size $h$ is given in Equation \ref{eq:concatcell} where $\sigma$ is the activation function, $w_t$ is the character embedding, $h_{t-1}$ is the previous hidden state, and $\mathbf{W} \in \mathbb{R}^{e + h \times h}$ and $b \in \mathbb{R}^{h}$ are the recurrent layer weight matrix and bias vector.
\begin{equation}
\label{eq:concatcell}
h_t = \sigma([w_t, h_{t-1}]\mathbf{W} + b + \mathbf{V}u)
\end{equation}

Adapting just the bias vector is a significant limitation. The FactorCell model, \cite{jaech2017low}, remedies this by letting the user embedding transform the weights of the recurrent layer via the use of a low-rank adaptation matrix. The FactorCell uses a weight matrix $\mathbf{W}'= \mathbf{W} + \mathbf{A}$ that has been additively transformed by a personalized low-rank matrix $\mathbf{A}$.  Because the FactorCell weight matrix $\mathbf{W}'$ is different for each user (See Equation \ref{eq:factorcell}), it allows for a much stronger adaptation than what is possible using the more standard ConcatCell model.\footnote{In the case of an LSTM, $\mathbf{W}'$ is extended to incorporate all of the gates.}
\begin{equation}
\label{eq:factorcell}
h_t = \sigma([w_t, h_{t-1}]\mathbf{W}' + b)
\end{equation}

The low-rank adaptation matrix $\mathbf{A}$ is generated by taking the product between a user's $m$ dimensional embedding and left and right bases tensors, $\mathbf{Z}_L \in \mathbb{R}^{m \times e + h \times r}$ and $\mathbf{Z}_R \in \mathbb{R}^{r \times h \times m}$ as so, 
\begin{equation}
\mathbf{A} = (u_i \times_1 \mathbf{Z}_L)(\mathbf{Z}_R \times_3 u_i)
\end{equation}
where $\times_i$ denotes the mode-i tensor product. The above product selects a user specific adaptation matrix by taking a weighted combination of the $m$ rank $r$ matrices held between $\mathbf{Z}_L$ and $\mathbf{Z}_R$. The rank, $r$, is a hyperparameter which controls the degree of personalization.

\section{Data}
\label{sec:data}

Our experiments make use of the AOL Query data collected over three months in 2006 \cite{pass2006picture}. The first six of the ten files were used for training. This contains approximately 12 million queries from 173,000 users for an average of 70 queries per user (median 15). A set of 240,000 queries from those same users (2\% of the data) was reserved for tuning and validation. From the remaining files, one million queries from 30,000 users are used to test the models on a disjoint set of users. 

\section{Experiments}
\label{sec:experiments}

\subsection{Implementation Details}

The vocabulary consists of 79 characters including special start and stop  tokens. Models were trained for six epochs. The Adam optimizer is used during training with a learning rate of ${10}^{-3}$ \cite{kingma2014adam}. When updating the user embeddings during evaluation, we found that it is easier to use an optimizer without momentum. We use Adadelta \cite{adadelta} and tune the online learning rate to give the best perplexity on a held-out set of 12,000 queries, having previously verified that perplexity is a good indicator of performance on the QAC task.\footnote{Code at http://github.com/ajaech/query\_completion}

The language model is a single-layer character-level LSTM with coupled input and forget gates and layer normalization \cite{melis2017state,ba2016layer}. We do experiments on two model configurations: small and large. The small models use an LSTM hidden state size of 300 and 20 dimensional user embeddings. The large models use a hidden state size of 600 and 40 dimensional user embeddings. Both sizes use 24 dimensional character embeddings. For the small sized models, we experimented with different values of the FactorCell rank hyperparameter between 30 and 50 dimensions finding that bigger rank is better. The large sized models used a fixed value of 60 for the rank hyperparemeter. During training only and due to limited computational resources, queries are truncated to a length of 40 characters. 

Prefixes are selected uniformly at random with the constraint that they contain at least two characters in the prefix and that there is at least one character in the completion. To generate completions using beam search, we use a beam width of 100 and a branching factor of 4. Results are reported using mean reciprocal rank (MRR), the standard method of evaluating QAC systems. It is the mean of the reciprocal rank of the true completion in the top ten proposed completions. The reciprocal rank is zero if the true completion is not in the top ten.

Neural models are compared against an MPC baseline. Following \newcite{park2017neural}, we remove queries seen less than three times from the MPC training data.

\subsection{Results}

Table \ref{table:unseen_users} compares the performance of the different models against the MPC baseline on a test set of one million queries from a user population that is disjoint with the training set. Results are presented separately for prefixes that are seen or unseen in the training data. Consistent with prior work, the neural models do better than the MPC baseline. The personalized models are both better than the unadapted one. The FactorCell model is the best overall in both the big and small sized experiments, but the gain is mainly for the seen prefixes.

\begin{table}[]
\centering
\begin{tabular}{ccrrr}
\textbf{Size} & \textbf{Model}  & \multicolumn{1}{c}{\textbf{Seen}}   & \multicolumn{1}{c}{\textbf{Unseen}} & \multicolumn{1}{c}{\textbf{All}} \\ \hline
& MPC          & .292  & .000  & .203\\ \hline
 & Unadapted   &   .292   &   .256   &   .267 \\
\textbf{(S)} & ConcatCell   &   .296   &   .263   &   .273 \\
 & FactorCell   &   .300   &   .264   &   .275 \\\hline
 & Unadapted   &   .324   &   .286   &   .297 \\
\textbf{(B)} & ConcatCell   &   .330   &   .298   &   .308 \\
 & FactorCell   &   .335   &   .298   &   .309 \\ \hline
\end{tabular}
\caption{MRR reported for seen and unseen prefixes for small (S) and big (B) models.}
\label{table:unseen_users}
\end{table}

\begin{figure}
    \centering
    \includegraphics[width=0.45\textwidth]{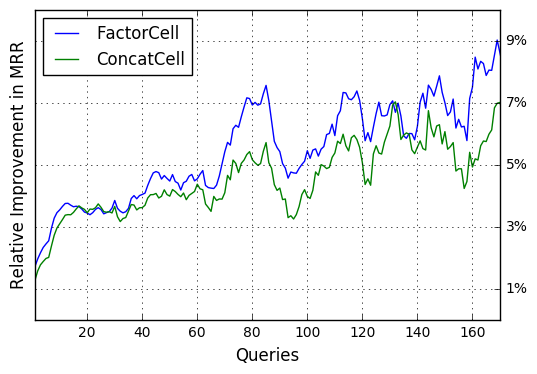}
    \caption{Relative improvement in MRR over the unpersonalized model versus queries seen using the large size models. Plot uses a moving average of width 9 to reduce noise.}
    \label{fig:improvement}
\end{figure}

Figure \ref{fig:improvement} shows the relative improvement in MRR over an unpersonalized model versus the number of queries seen per user. Both the FactorCell and the ConcatCell show continued improvement as more queries from each user are seen, and the FactorCell outperforms the ConcatCell by an increasing margin over time. In the long run, we expect that the system will have seen many queries from most users. Therefore, the right side of Figure \ref{fig:improvement}, where the relative gain of FactorCell is up to 2\% better than that of the ConcatCell, is more indicative of the potential of these models for active users.
Since the data was collected over a limited time frame and half of all users have fifteen or fewer queries, the results in Table \ref{table:unseen_users} do not reflect the full benefit of personalization.

\begin{figure}
    \centering
    \includegraphics[width=0.4\textwidth]{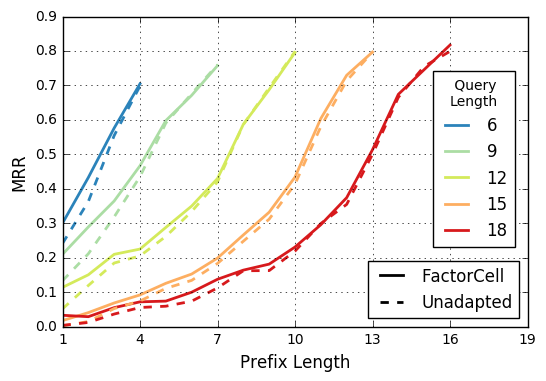}
    \caption{MRR by prefix and query lengths for the large FactorCell and unadapted models with the first 50 queries per user excluded.}
    \label{fig:prefix_len}
\end{figure}

Figure \ref{fig:prefix_len} shows the MRR for different prefix and query lengths. We find that longer prefixes help the model make longer completions and (more obviously) shorter completions have higher MRR. Comparing the personalized model against the unpersonalized baseline, we see that the biggest gains are for short queries and prefixes of length one or two.

We found that one reason why the FactorCell outperforms the ConcatCell is that it is able to pick up sooner on the repetitive search behaviors that some users have. This commonly happens for navigational queries where someone searches for the name of their favorite website once or more per day. At the extreme tail there are users who search for nothing but free online poker. Both models do well on these highly predictable users but the FactorCell is generally a bit quicker to adapt.

We conducted case studies to better understand what information is represented in the user embeddings and what makes the FactorCell different from the ConcatCell. From a cold start user embedding we ran two queries and allowed the model to update the user embedding. Then, we ranked the most frequent 1,500 queries based on the ratio of their likelihood from before and after updating the user embeddings. 

Tables \ref{table:teen_queries} and \ref{table:high_fashion} show the queries with the highest relative likelihood of the adapted vs.\ unadapted models after two related search queries:
``high school softball" and ``math homework help" for Table \ref{table:teen_queries}, and ``Prada handbags" and ``Versace eyewear" for Table \ref{table:high_fashion}. In both cases, the FactorCell model examples are more semantically coherent than the ConcatCell examples. In the first case, the
FactorCell model identifies queries that a high school student might make, including entertainment sources and a celebrity entertainer popular with that demographic.
In the second case, the FactorCell model chooses retailers that carry woman's apparel and those that sell home goods. While these companies' brands are not as luxurious as Prada or Versace, most of the top luxury brand names do not appear in the top 1,500 queries and our model may not be capable of being that specific.
There is no obvious semantic connection between the highest likelihood ratio phrases for the ConcatCell; it seems to be focusing more on orthography than semantics (e.g. ``home'' in the first example)..
Not shown are the queries which experienced the greatest decrease in likelihood. For the ``high school" case, these included searches for travel agencies and airline tickets---websites not targeted towards the high school age demographic. 


\begin{table}[]
\centering
\begin{tabular}{ccc}
\textbf{} & \textbf{FactorCell} & \textbf{ConcatCell} \\ \hline
1         & high school musical & horoscope           \\
2         & chris brown         & high school musical \\
3         & funnyjunk.com       & homes for sale      \\
4         & funbrain.com        & modular homes       \\
5         & chat room           & hair styles      
\end{tabular}
\caption{The five queries that have the greatest adapted vs.\ unadapted likelihood ratio
after searching for ``high school softball" and ``math homework help".}
\label{table:teen_queries}
\end{table}


\begin{table}[]
\centering
\begin{tabular}{ccc}
\textbf{} & \textbf{FactorCell} & \textbf{ConcatCell} \\ \hline
1         & neiman marcus       & craigslist nyc      \\
2         & pottery barn        & myspace layouts     \\
3         & jc penney           & verizon wireless    \\
4         & verizon wireless    & jensen ackles       \\
5         & bed bath and beyond & webster dictionary 
\end{tabular}
\caption{The five queries that have the greatest  adapted vs.\ unadapted likelihood ratio after searching for ``prada handbags" and ``versace eyewear".}
\label{table:high_fashion}
\end{table}

\section{Related Work}

While the standard implementation of MPC can not handle unseen prefixes, there are variants which do have that ability. \newcite{park2017neural} find that the neural LM outperforms MPC even when MPC has been augmented with the approach from \newcite{mitra2015query} for handling rare prefixes. There has also been work on personalizing MPC \cite{shokouhi2013learning,cai2014time}. We did not compare against these specific models because our goal was to show how personalization can improve the already-proven generative neural model approach. RNN's have also previously been used for the related task of next query suggestion \cite{sordoni2015hierarchical}.

Our results are not directly comparable to \newcite{park2017neural} or \newcite{mitra2015query} due to differences in the partitioning of the data and the method for selecting random prefixes. Prior work partitions the data by time instead of by user. Splitting by users is necessary in order to properly test personalization over longer time ranges.

\newcite{anonymous2018realtime} show how spelling correction can be integrated into an RNN language model query auto-completion system and how the completions can be generated in real time using a GPU. Our method of updating the model during evaluation resembles work on dynamic evaluation for language modeling \cite{krause2017dynamic}, but differs in that only the user embeddings (latent demographic factors) are updated.

\section{Conclusion and Future Work}

Our experiments show that the LSTM model can be improved using personalization. The method of adapting the recurrent layer clearly matters and we obtained an advantage by using the FactorCell model. The reason the FactorCell does better is in part attributable to having two to three times as many parameters in the recurrent layer as either the ConcatCell or the unadapted models. By design, the adapted weight matrix $\mathbf{W}'$ only needs to be computed at most once per query and is reused many thousands of times during beam search. As a result, for a given latency budget, the FactorCell model outperforms the \newcite{mikolov2012context} model for LSTM adaptation. 

The cost for updating the user embeddings is similar to the cost of the forward pass and depends on the size of the user embedding, hidden state size, FactorCell rank, and query length. In most cases there will be time between queries for updates, but updates can be less frequent to reduce computational costs.

We also showed that language model personalization can be effective even on users who are not seen during training. The benefits of personalization are immediate and increase over time as the system continues to leverage the incoming data to build better user representations. The approach can easily be extended to include time as an additional conditioning factor. We leave the question of whether the results can be improved by combining the language model with MPC for future work.


\bibliography{naaclhlt2018}
\bibliographystyle{acl_natbib}

\end{document}